\documentclass[runningheads]{llncs}
\usepackage{graphicx}

\usepackage{tikz}
\usepackage{comment}
\usepackage{amsmath,amssymb} 
\usepackage{color}
\usepackage{orcidlink}
\usepackage[accsupp]{axessibility}  

\begin{document}
\pagestyle{headings}
\mainmatter
\def\ECCVSubNumber{4249}  

\title{Human-centric Image Cropping with Partition-aware and Content-preserving Features} 

\titlerunning{Human-centric Image Cropping}
%
\author{Bo Zhang\orcidlink{0000-0001-6577-7484} \and
Li Niu\thanks{Corresponding author}\orcidlink{0000-0003-1970-8634} \and
Xing Zhao\orcidlink{0000-0001-9234-3235} \and
Liqing Zhang\orcidlink{0000-0001-7597-8503}}
\authorrunning{B{.} Zhang et al.}
%
\institute{MoE Key Lab of Artificial Intelligence, Shanghai Jiao Tong University, China \\ \email{\{bo-zhang,ustcnewly,1033874657\}@sjtu.edu.cn, zhang-lq@cs.sjtu.edu.cn}
}
\maketitle

\begin{abstract}
Image cropping aims to find visually appealing crops in an image, which is an important yet challenging task. In this paper, we consider a specific and practical application: human-centric image cropping, which focuses on the depiction of a person. 
To this end, we propose a human-centric image cropping method with two novel feature designs for the candidate crop: partition-aware feature and content-preserving feature. For partition-aware feature, we divide the whole image into nine partitions based on the human bounding box and treat different partitions in a candidate crop differently conditioned on the human information. For content-preserving feature, we predict a heatmap indicating the important content to be included in a good crop, and extract the geometric relation between the heatmap and a candidate crop. Extensive experiments demonstrate that our method can perform favorably against state-of-the-art image cropping methods on human-centric image cropping task. Code is available at \textcolor{blue}{\url{https://github.com/bcmi/Human-Centric-Image-Cropping}}.
\end{abstract}

\section{Introduction}
\label{sec:introduction}

Image cropping aims to automatically find visually appealing crops in an image, which is critical in various down-stream applications, \eg, photo post-processing \cite{chen2017learning}, view recommendation \cite{li2019fast,wei2018good,li2018a2}, image thumbnailing \cite{esmaeili2017fast,chen2018cropnet}, and camera view adjustment suggestion \cite{su2021camera}. 
In this paper, we address a specific and practical application: human-centric image cropping, which focuses on the depiction of a person and benefits a variety of applications, including portrait enhancement \cite{zhang2005auto} and portrait composition assistance \cite{zhang2018pose,zhang2012aesthetic}.
For a human-centric image, a good crop depends on the position of the human in the crop, human information, and the content of interest, which makes human-centric image cropping challenging.

Several previous works \cite{zhang2018pose,cavalcanti2010combining,zhang2005auto} have already focused on portrait photograph cropping, which extracted hand-crafted features from the results of saliency detection, human face detection, or human pose estimation. 
However, extracting hand-crafted features is laborious and the hand-crafted features are generally not robust for modeling the huge aesthetic space \cite{deng2017image}.
Recently, numerous methods \cite{hong2021composing,zeng2020cropping,Tu2020ImageCW,li2020composing,li2020learning,wang2018deep} addressed image cropping task in a data-driven manner, in which models are directly trained with the human-annotated datasets \cite{chen2017quantitative,zeng2019reliable,wei2018good}.
However, for human-centric images, these methods rarely explicitly consider human information. In contrast, we show that exploiting human information can significantly help  obtain good crops.
Based on the general pipeline of data-driven methods, we propose two innovations for human-centric image cropping.

\begin{figure}[tbp]
    \centering
    \includegraphics[width=0.6\linewidth]{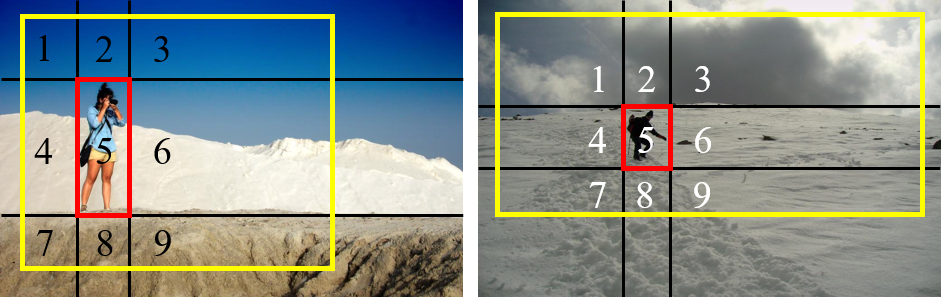}
    \caption{Illustration of the motivation behind partition-aware feature.  The whole image is divided into nine partitions based on the human bounding box (red). To produce the best crop (yellow), the aesthetic contribution of similar content in different partitions depends on its relative position to human subject}
    \label{fig:ill_partition_aware}
\end{figure}

In this paper, we refer to the images that meet the following conditions as human-centric images: 1) The image subject is single person, while there can be other people in the background. 2) The area of the human bounding box does not exceed 90\% of the entire frame.
Given a human-centric image, the whole image can be divided into nine partitions based on the human bounding box (see Figure \ref{fig:ill_partition_aware}). Generally, the aesthetic contribution of similar content in different partitions depends on its relative position to the human subject. 
For example, in Figure \ref{fig:ill_partition_aware}, partitions 4 and 6 in the left subfigure have similar content, but the best crop preserves more content in partition 6 because the person looks to the right, making the content in partition 6 visually more important \cite{Freeman2007ThePE}. 
Similarly, partition 4 and partition 8 in the right subfigure also have similar content, but the best crop preserves more content in partition 4, probably because the person is moving forward and the content behind him becomes less important. 
Therefore, when extracting features of candidate crops for aesthetic evaluation, we should consider the partition location and human information (\eg, human posture, face orientation).
To this end, we propose a novel partition-aware feature by incorporating partition and human information, which enables treating different partitions in a candidate crop differently conditioned on the human information.

\begin{figure}[tbp]
    \centering
    \includegraphics[width=0.6\linewidth]{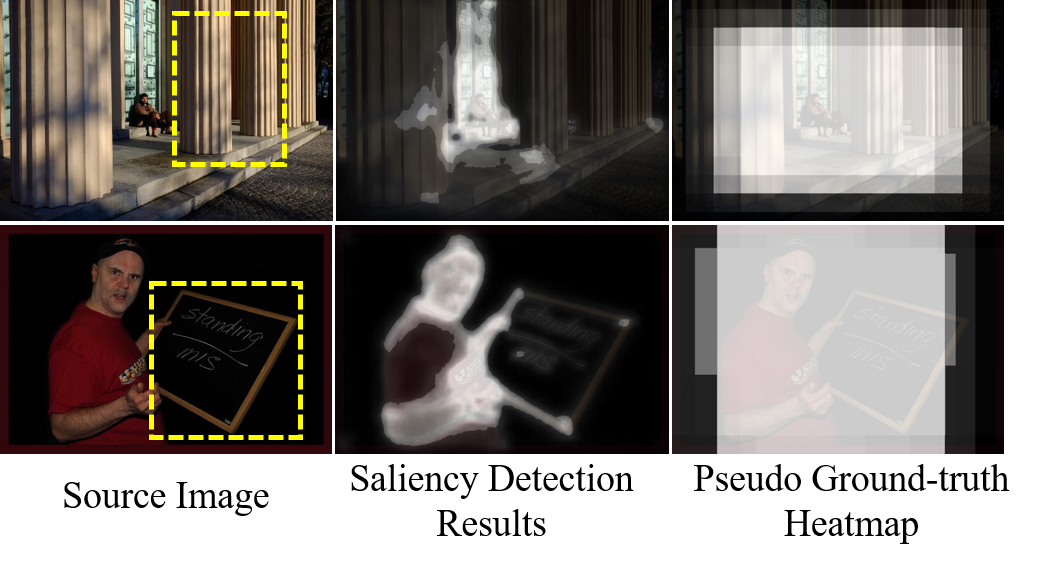}
    \caption{The comparison between saliency detection \cite{goferman2012context} and the pseudo ground-truth heatmap of important content (see Section \ref{sec:content_preserve}). The important content of human-centric images may contain interesting objects (\eg, the landmark in the top row) and the objects that person interacts with (\eg, the blackboard in the bottom row)}
    \label{fig:ill_important_content}
\end{figure}

Furthermore, a good crop should preserve the important content of source image \cite{fang2014automatic}, which is dubbed as ``content-preserving". However, to the best of our knowledge, there is no image cropping dataset that provides the annotation of important content.
Existing methods \cite{li2019image,saliency-wang2017deep,chen2016automatic,fang2014automatic} determine important content mainly based on their visual saliency by assuming that the most salient object is the most important content.
In human-centric images, important content may imply key human parts (\eg, face, hands), interesting objects (\eg, landmark), and the objects (\eg, racket, bicycle) that person interacts with.
However, as shown in Figure \ref{fig:ill_important_content}, saliency may not capture the abovementioned objects very well.
Here we adopt an unsupervised saliency detection method \cite{goferman2012context}. We have also tried several supervised methods \cite{hou2017deeply,zhao2019pyramid,chen2020global}, which proves to be less effective. This is probably because that the unsupervised method has no dependence on training data and generalizes better on the image cropping datasets.  
Given an image with multiple annotated candidate crops \cite{zeng2019reliable,wei2018good}, we conjecture that the candidate crops with relatively high scores are more likely to contain important objects. Thus, we use highly scored crops to produce pseudo ground-truth heatmap of important content (see Figure \ref{fig:ill_important_content}), which is used to supervise the heatmap prediction.  
Additionally, previous content-preserving methods \cite{ardizzone2013saliency,chen2016automatic} typically designed a hand-crafted algorithm based on certain principles (\eg, maintaining the most salient region).
Differently, we automatically learn content-preserving feature to capture the geometric relation between the predicted heatmap and each candidate crop, which represents how well each candidate crop preserves the important content. 

Finally, for each candidate crop, we extract its partition-aware feature and content-preserving feature to predict an aesthetic score.
The main contributions of this paper can be summarized as follows: 
1) We propose a novel partition-aware feature to improve human-centric image cropping by exploiting human information, which allows to treat different regions in a candidate crop differently conditioned on the human information.
2) We design a novel approach to locate important content and a novel content-preserving feature to characterize the preservation of important content in a candidate crop.
3) We demonstrate that our model outperforms the state-of-the-art image cropping methods on the human-centric images of several benchmark datasets.

\section{Related Work}
\label{sec:related_work}
Following \cite{zeng2020cropping,li2020composing}, we divide existing image cropping methods into three categories according to the criteria for evaluating candidate crops, \ie, attention-guided, aesthetics-informed, and data-driven.

\noindent\textit{Attention-guided Image Cropping:} Attention-guided methods \cite{lu2019end,li2019image,li2019collaborative,chen2016automatic,fang2014automatic,saliency-sun2013scale,zhang2005auto} assumed that the best crops should preserve visually important content, which is usually determined by the saliency detection methods \cite{vig2014large,hou2007saliency}. 
Usually, the view with the highest average saliency score is selected as the best crop. However, saliency may not accurately reflect the content of interest for human-centric images (see Figure \ref{fig:ill_important_content}). 
Differently, we assume that the content that appears in multiple highly scored crops is more likely to be important content, leading to more flexible and practical important content estimation.   

\noindent\textit{Aesthetics-informed Image Cropping:} The aesthetics-informed methods evaluated candidates by comparing the overall aesthetic quality of different crops. To achieve this, earlier methods \cite{zhang2013weakly,yan2013learning,zhang2012probabilistic} usually employed hand-crafted features or composition rules. However, the simple hand-crafted features may not accurately predict the complicated image aesthetics \cite{zeng2020cropping}.

\noindent\textit{Data-driven Image Cropping:} Most recent methods address the task in a data-driven manner. Some methods \cite{chen2017learning,kao2017automatic,MNA-CNN-mai2016composition} trained a general aesthetic evaluator on image aesthetic datasets to facilitate image cropping.
With the aid of image cropping datasets \cite{chen2017quantitative,zeng2019reliable,wei2018good}, numerous methods \cite{hong2021composing,zeng2019reliable,Tu2020ImageCW,lu2019listwise,li2020composing,wei2018good} used pairwise learning to train an end-to-end model on these datasets, which can generate crop-level scores for ranking different candidate crops.

Our method is developed based on the general pipeline of the data-driven methods, but is specially tailored to human-centric image cropping with two innovations, \ie, partition-aware and content-preserving features.

\section{Methodology}
\label{sec:methodology}

\subsection{Overview}
\label{sec:overview}
\begin{figure*}
    \centering
    \includegraphics[width=1\linewidth]{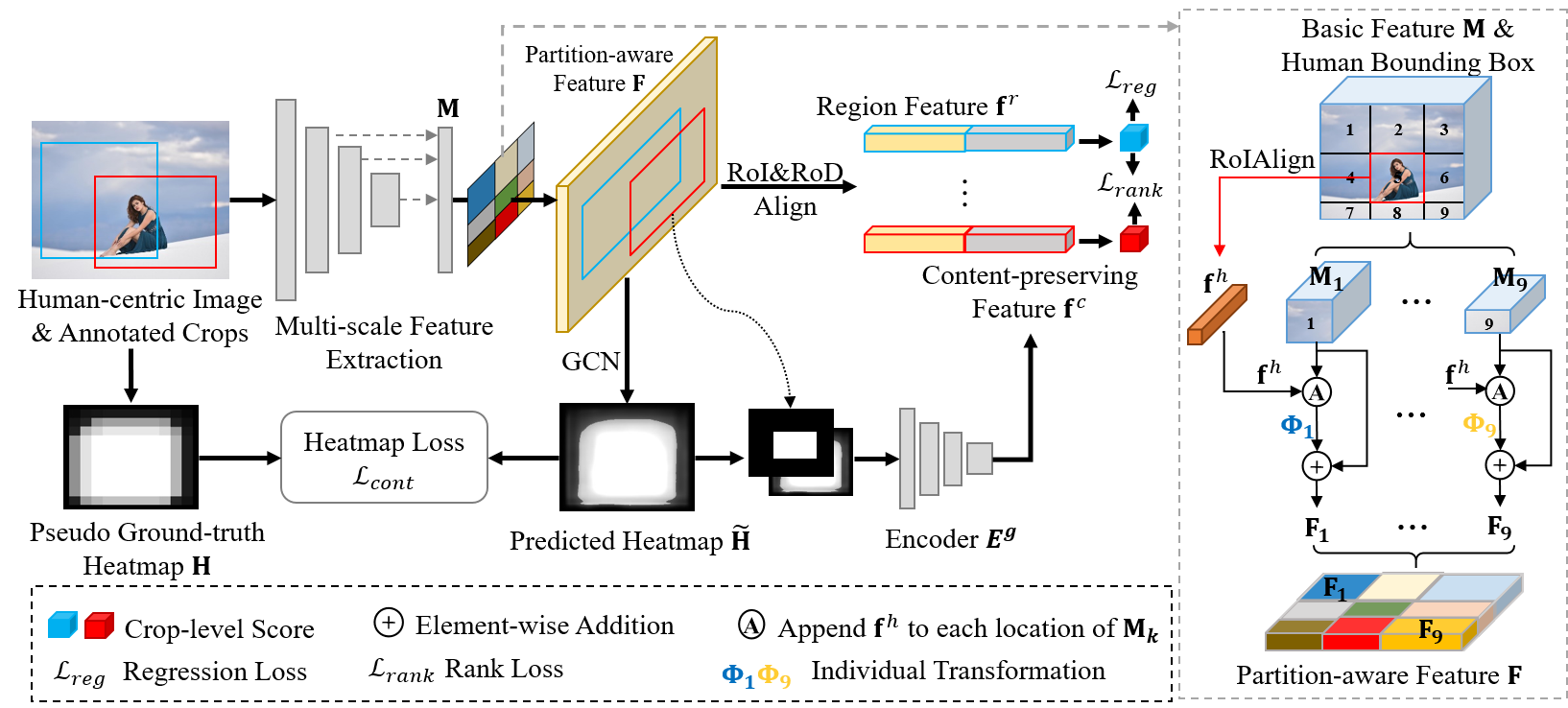}
    \caption{The flowchart of our method for human-centric image cropping (left) and the proposed partition-aware feature (right). We use pretrained VGG16 \cite{vgg-simonyan2014very} as backbone to extract basic feature map $\mathbf{M}$, from which we derive partition-aware feature map $\mathbf{F}$. Besides, we use the region feature obtained by RoIAlign \cite{he2017mask} and RoDAlign \cite{zeng2019reliable} schemes, and content-preserving feature to predict scores for each candidate crop}
    \label{fig:pipeline}
\end{figure*}

The flowchart of the proposed method is illustrated in Figure \ref{fig:pipeline}, in which we adopt a similar pipeline as \cite{zeng2020cropping,li2020composing}.
Given an image, we first integrate multi-scale feature maps from a pretrained backbone (\eg, VGG16 \cite{vgg-simonyan2014very}) to obtain the basic feature map. After that, we update the basic feature map to the partition-aware feature map, based on which we extract partition-aware region feature and content-preserving feature for each candidate crop. Finally, we predict the crop-level score using the concatenation of partition-aware region feature and content-preserving feature. 

\subsection{Partition-aware Feature}
\label{sec:partition_aware}
To acquire the human bounding box, we leverage Faster R-CNN~\cite{ren2015faster} trained on Visual Genome~\cite{krishna2017visual} to detect human subjects for human-centric images. We check each image to ensure that the predicted bounding box correctly encloses the main human subject. We describe how to determine the main subject and discuss the robustness of our method against human detection in Supplementary.
As illustrated in Figure \ref{fig:ill_partition_aware}, the whole image can be divided into nine non-overlapping partitions by the human bounding box. We conjecture that the aesthetic value of similar content in different partitions often varies with its relative position to the human subject, so the feature map should be partition-aware.

To achieve this goal, we derive partition-aware feature map from the basic feature map, as illustrated in the left subfigure in Figure \ref{fig:pipeline}. Given an $H \times W$ basic feature map $\mathbf{M}$ with $C$ channels, we partition it into nine regions $\{\mathbf{M}_{k}|_{k=1}^9\}$ with $\mathbf{M}_{k}$ being the $k$-th partition. Considering that the relative position of each partition to human subject is conditioned on the human information (\eg, face orientation and posture), as exemplified in Figure \ref{fig:ill_partition_aware}, we extract human feature $\mathbf{f}^h$ using RoIAlign \cite{he2017mask} to implicitly encode the aforementioned human information and reduce its dimension to $\frac{C}{2}$, which is appended to each location of the basic feature map. The resultant feature map is represented by $\hat{\mathbf{M}}$ with $\hat{\mathbf{M}}_{k}$ being the $k$-th partition. To explicitly tackle different partitions differently, we employ nine individual nonlinear transformations for each partition to update their features with residual learning by
\begin{equation}
    \mathbf{F}_k = \Phi_k(\hat{\mathbf{M}}_k) + \mathbf{M}_k,
\label{eq:partition_aware}
\end{equation}
where we use a $3 \times 3$ convolutional (conv) layer with $C$ output channels followed by $\mathrm{ReLU}$ as the transformation function $\Phi_{k}(\cdot)$.

After that, we obtain a new feature map $\mathbf{F}$ by combining all updated partitions $\{\mathbf{F}_{k}|_{k=1}^9\}$, which has the same size as the basic feature map $\mathbf{M}$. \emph{By integrating human feature and employing partition-specific transformations, similar contents in different partitions can produce different responses conditioned on the human information.} Thus, we refer to $\mathbf{F}$ as partition-aware feature map. Following \cite{zeng2019reliable,li2020composing}, given a candidate crop, we employ RoIAlign \cite{he2017mask} and RoDAlign \cite{zeng2019reliable} schemes to extract its partition-aware region feature based on $\mathbf{F}$, denoted as $\mathbf{f}^r$.

\subsection{Content-preserving Feature}
\label{sec:content_preserve}
Apart from the position of the human in the crop and human information, the preservation of important content also plays an important role when evaluating candidate crops. So we propose to predict a heatmap to indicate the location of important content and then automatically learn content-preserving features to augment our method. 

\noindent\textit{Graph-based Region Relation Mining:} Considering that important content estimation may benefit from exploiting the mutual relation between different regions, we construct a graph over the partition-aware feature map $\mathbf{F}$ and apply graph convolution \cite{scarselli2008graph,Li2018DeeperII}. Specifically, we reshape the partition-aware feature maps into a matrix $\mathbf{\bar{F}} \in \mathcal{R}^{L \times C}$, where $L = H \times W$. Each pixel-wise feature vector in $\mathbf{\bar{F}}$ is a graph node that represents one local region in the image. 

To model the relation between pairwise regions, we define the adjacency matrix $\mathbf{A}  \in \mathcal{R}^{L \times L}$ according to the cosine similarity of region features following \cite{she2021hierarchical}. Then we perform reasoning over the graph $\mathbf{\bar{F}}$ by graph convolution \cite{scarselli2008graph}:
\begin{equation}
    \mathbf{\bar{F}}' = \sigma(\mathbf{A} \mathbf{\bar{F}} \mathbf{\Theta}),
\end{equation}
where $\mathbf{\Theta} \in \mathcal{R}^{C \times C}$ is the the trainable weight matrix of the graph convolution layer and $\sigma(\cdot)$ is  $\mathrm{ReLU}$ activation. Then, we reshape $\mathbf{\bar{F}}'$ back to  $\mathbf{F}'\in \mathcal{R}^{H \times W \times C}$. 
Compared with the conventional convolution, graph convolution allows the message flow across local regions, which is helpful for important content prediction (see Section \ref{sec:analysis_heatmap}).

\noindent\textit{Important Content Estimation:} To obtain a high-resolution feature map for fine-grained important content localization, we upsample $\mathbf{F}'$ by four times followed by $3 \times 3$ conv and $\mathrm{ReLU}$.
Based on the upsampled feature map, we apply a prediction head (\ie, $1 \times 1$ conv followed by $\mathrm{Sigmoid}$ function) to produce a heatmap $\widetilde{\mathbf{H}}$ in the range of $[0,1]$, in which larger score indicates more important content.

There is no ground-truth heatmap for important content. Nevertheless, existing cropping datasets \cite{zeng2019reliable,wei2018good} are associated with multiple scored crops for each image, with a larger score indicating higher aesthetic quality. \emph{Based on the assumption that highly scored crops are more likely to contain important content, we propose to generate pseudo ground-truth heatmap from the weighted average of highly scored crops.}

Specifically, given an image with multiple candidate crops, we suppose the score of the $m$-th crop to be $y_{m}$.    
We take the average score of all crops in the dataset as the threshold to select highly scored crops, and convert the bounding box of each selected crop to a binary map, which is resized to the same size as the predicted heatmap $\widetilde{\mathbf{H}}$. We obtain the pseudo ground-truth heatmap $\mathbf{H}$ via the weighted average of all binary maps, in which larger weight is assigned to the crop with higher score. Here, we perform $\mathrm{softmax}$ normalization to produce the weights for each highly scored crop: $\omega_{m}=\frac{\mathrm{exp}(y_{m})}{\sum^{N_h}_{m=1} \mathrm{exp}(y_{m})}$, in which $N_h$ is the number of highly scored crops. In the training stage, we employ an L1 loss to supervise the heatmap learning:
\begin{equation}
    \mathcal{L}_{cont} = \|\mathbf{H}- \widetilde{\mathbf{H}} \|_1.
\label{eq:content_loss}
\end{equation}

As demonstrated in Figure \ref{fig:ill_important_content}, the pseudo ground-truth heatmaps can highlight the attractive objects in the background and the objects that person interacts with, which are often ignored by saliency detection methods.   

\noindent\textit{Content-preserving Feature Extraction:} To leverage the prior information of important content \cite{wei2018good}, previous content-preserving methods typically relied on certain heuristic principles (\eg, minimizing the cropping area while maximizing its attention value \cite{chen2016automatic}), which require extensive manual designs.  
\emph{Instead, we guide the network to automatically learn how well each candidate crop preserves the important content. The idea is learning a content-preserving feature to capture the geometric relation between the heatmap and each candidate crop.} Specifically, for the $m$-th candidate crop, we concatenate its corresponding binary map and the predicted heatmap $\widetilde{\mathbf{H}}$ channel-wisely. Then, we apply an encoder $E^{g}$ to the concatenated maps to extract the content-preserving feature $\mathbf{f}^{c}$. 

Finally, for each candidate crop, we concatenate its partition-aware region feature  $\mathbf{f}^r$ and content-preserving feature $\mathbf{f}^{c}$, which is passed through a fully connected (fc) layer to get the aesthetic score for this crop.

\subsection{Network Optimization}
\label{sec:network_optimization}
We train the proposed model with a multi-task loss function in an end-to-end manner. Given an image containing $N$ candidate crops, the ground-truth and predicted scores of the $m$-th crop are denoted by $y_{m}$ and $\tilde{y}_{m}$, respectively. We first employ a smooth L1 loss \cite{ren2015faster} for the score regression considering its robustness to outliers:
\begin{equation}
    \mathcal{L}_{reg} = \frac{1}{N} \sum^{N}_{m=1} \mathcal{L}_{s1}(y_{m} - \tilde{y}_{m}),
\end{equation}
where $\mathcal{L}_{s1}(\cdot)$ represents the smooth L1 loss:
\begin{equation}
    \mathcal{L}_{s1}(x)=\left\{\begin{array}{cc}
                0.5 x^{2}, & \mathrm{if}\ |x| <1, \\
                |x|-0.5, &   \mathrm{otherwise}.
                \end{array}\right.
\end{equation}
Besides regression loss, we also use a ranking loss \cite{li2020composing} to learn the relative ranking order between pairwise crops explicitly, which is beneficial for enhancing the ability of ranking crops with similar content. With $e_{m,n} = y_m - y_n$ and $\tilde{e}_{m,n} = \tilde{y}_m - \tilde{y}_n$, the ranking loss is computed by
\begin{equation}
    \mathcal{L}_{rank}=\frac{\sum_{m,n} \max \big(0,\mathrm{sign}(e_{m,n})(e_{m,n} - \tilde{e}_{m,n}) \big) }{N(N-1) / 2}.
\end{equation}

After including the heatmap prediction loss in Eqn.(\ref{eq:content_loss}), the total loss is summarized as
\begin{equation}
    \mathcal{L} = \mathcal{L}_{reg} + \mathcal{L}_{rank} + \lambda \mathcal{L}_{cont},
\label{eq:total_loss}
\end{equation}
in which the trade-off parameter $\lambda$ is set as 1 via cross-validation (see Supplementary).

Limited by the small number of annotated human-centric images in existing image cropping datasets \cite{zeng2019reliable,wei2018good}, only training on human-centric images would lead to weak generalization ability to the test set. Therefore, \emph{we employ both human-centric and non-human-centric images to train our model.} 
For non-human-centric images, we use the basic feature map $\mathbf{M}$ to replace the partition-aware feature map $\mathbf{F}$, because there could be no dominant subject used to partition the image. Besides, we extract content-preserving features from non-human-centric images in the same way as human-centric images, because preserving important content is also crucial for non-human-centric images. 
In this way, our model is able to train and infer on both human-centric images and non-human-centric images.

\section{Experiments}
\label{sec:experiment}

\subsection{Datasets and Evaluation Metrics}
\label{sec:dataset_and_metrics}
We conduct experiments on the recent GAICD dataset \cite{zeng2019reliable}, which contains 1,236 images (1,036 for training and 200 for testing). Each image has an average of 86 annotated crops. 
There are 339 and 50 human-centric images in the training set and test set of GAICD dataset, respectively. As described in Section \ref{sec:network_optimization}, we employ the whole training set (1,036 images) for training and evaluate on the human-centric samples of the test set.
Following \cite{zeng2019reliable}, three evaluation metrics are employed in our experiments, including the average Spearman's rank-order correlation coefficient ($\overline{SRCC}$) and averaged top-$N$ accuracy ($\overline{Acc_{N}}$) for both $N=5$ and $N=10$. The $\overline{SRCC}$ computes the rank correlation between the ground-truth and predicted scores of crops for each image, which is used to evaluate the ability of correctly ranking multiple annotated crops. The $\overline{Acc_N}$ measures the ability to return the best crops.

Apart from GAICD dataset, we also collect 176 and 39 human-centric images from existing FCDB \cite{chen2017quantitative} and FLMS \cite{fang2014automatic} datasets, respectively. 
We evaluate our method on the collected 215 human-centric images from two datasets, following the experimental setting in \cite{lu2019listwise,Tu2020ImageCW}: training the model on CPC dataset \cite{wei2018good}, and using intersection of union (IoU) and boundary displacement (Disp) for performance evaluation.
Note that we train on the whole CPC dataset, which contains 10,797 images including 1,154 human-centric images and each image has 24 annotated crops.

\begin{table}[t]
    \caption{Ablation studies of the proposed method. $\mathbf{f}^{h}$: human feature. ``res'': update partition-aware feature with residual learning. $K$: number of partitions. ``conv'': replace GCN with standard convolution. $\widetilde{\mathbf{H}}$: predicted important content heatmap. ``saliency'': replace heatmap with saliency map \cite{goferman2012context}. $\mathbf{f}^c$: content-preserving feature}
    \label{table:ablation_study}
    \begin{center}
        \setlength{\tabcolsep}{3.5mm}
        \begin{tabular}{c|cc|ccc}
        \hline
          & Partition      & Content      & $\overline{SRCC}\uparrow$  & $\overline{Acc_5}\uparrow$ & $\overline{Acc_{10}}\uparrow$ \\
          \hline \hline
        1 &                              &                                 & 0.744             & 52.0              & 70.5               \\ \hline
        2 & \checkmark                   &                                 & 0.774             & 54.8              & 74.3               \\
        3 & w/o $\mathbf{f}^{h}$         &                                 & 0.769             & 54.2              & 73.8               \\
        4 & w/o res                      &                                 & 0.764             & 53.9              & 73.5               \\ 
        5 & $K=1$                        &                                 & 0.746             & 52.1              & 70.8               \\
        6 & $K=2$                        &                                 & 0.756             & 53.2              & 72.4               \\\hline
        7 &                              & \checkmark                      & 0.781             & 56.8              & 75.6               \\
        8 &                              & conv                            & 0.762            & 54.0               & 73.0               \\
        9 &                              & w/o $\widetilde{\mathbf{H}}$        & 0.741             & 50.9              & 69.5               \\
        10 &                             & saliency                        & 0.752             & 52.5              & 71.8               \\ 
        11 &                             & only $\mathbf{f}^c$             & 0.643             & 35.2              & 49.1               \\ \hline
        12 & \checkmark                  & \checkmark                      & \textbf{0.795}    & \textbf{59.7}     & \textbf{77.0}       \\ \hline
        \end{tabular}
    \end{center}
\end{table}

\subsection{Implementation Details}
\label{sec:implement_datails}

Following existing methods \cite{hong2021composing,Tu2020ImageCW,li2020composing}, we use VGG16 \cite{vgg-simonyan2014very} pretrained on ImageNet \cite{deng2009imagenet} as the backbone. We apply $1\times 1$ conv to unify the channel dimensions of the last three output feature maps as $256$ and add up three feature maps to produce the multi-scale feature map. We reduce the channel dimension of multi-scale feature map to 32 using a $1 \times 1$ conv, \ie, $C=32$. $E^{g}$ is implemented by two $3\times 3$ convs and pooling operations followed by a fc layer. The dimensions of partition-aware region feature $\mathbf{f}^{r}$ and content-preserving feature  $\mathbf{f}^{c}$ are both $256$. Similar to \cite{li2020composing,zeng2019reliable}, the short side of input images is resized to 256 and the aspect ratios remain unchanged.
We implement our method using PyTorch \cite{paszke2019pytorch} and set the random seed to 0. More implementation details can be found in Supplementary. 

\subsection{Ablation study}
\label{sec:ablation_study}
In this section, we start from the general pipeline of existing methods \cite{zeng2020cropping,li2020composing} and evaluate the effectiveness of two types of features. The results are summarized in Table \ref{table:ablation_study}. In the baseline (row 1), we only use the region feature extracted from the basic feature map $\mathbf{M}$ to predict scores for each crop.

\noindent\textit{Partition-aware Feature:} Based on row 1, we replace the region feature with our proposed partition-aware region feature $\mathbf{f}^r$ in row 2, which verifies the effectiveness of partition-aware feature. Next, we conduct ablation studies based on row 2. 
First, we remove the human feature $\mathbf{f}^h$ and observe performance drop, which corroborates the importance of conditional human information. In Eqn.(\ref{eq:partition_aware}), we adopt $\Phi_k(\cdot)$ to learn the residual. Based on row 2, we remove the residual strategy by using $\mathbf{F}_k = \Phi_k(\hat{\mathbf{M}}_k)$. The comparison between row 4 and row 2 demonstrates the benefit of residual learning.  Recall that each image is divided into nine partitions by the human bounding box. Based on row 2, we explore using one partition ($K=1$) and two partitions ($K=2$). When $K=1$, we apply the same transformation $\Phi(\cdot)$ to the whole image. When $K=2$, we divide the image into human bounding box and the outside region. By comparing $K=1,2,9$,  we observe that $K=9$ (row 2) achieves the best performance, because nine partitions can help capture more fine-grained partition-aware information.
Besides, we evaluate some direct ways to leverage human bounding box for image cropping, yet producing poor results (see Supplementary). 

\begin{table*}[t]
    \caption{Comparison with the state-of-the-art methods on human-centric images in GAICD \cite{zeng2019reliable} dataset. GAIC(ext) \cite{zeng2020cropping} is the extension of GAIC\cite{zeng2019reliable}. The results marked with * are obtained using the released models from original papers}
    \label{table:comparison_GAICD}
    \begin{center}
        \setlength{\tabcolsep}{2mm}
        \begin{tabular}{l|cc|ccc}
        \hline
        Method     & Backbone    & Training Data & $\overline{SRCC}\uparrow$  & $\overline{Acc_5}\uparrow$ & $\overline{Acc_{10}}\uparrow$ \\ \hline \hline
        VFN* \cite{chen2017learning}        & AlextNet    & Flickr       & 0.332 & 10.1  & 21.1   \\
        VFN \cite{chen2017learning}        & VGG16       & GAICD        & 0.648 & 41.3  & 60.2   \\
        VEN* \cite{wei2018good}             & VGG16       & CPC          & 0.641 & 22.4  & 36.2   \\
        VEN \cite{wei2018good}             & VGG16       & GAICD        & 0.683 & 50.1  & 65.1   \\
        ASM-Net \cite{Tu2020ImageCW}      & VGG16       & GAICD        & 0.680  & 44.8  & 64.5   \\ 
        LVRN* \cite {lu2019listwise}        & VGG16       & CPC          & 0.664 & 30.7  & 49.0   \\
        LVRN \cite {lu2019listwise}        & VGG16       & GAICD        & 0.716 & 44.8  & 66.0   \\
        GAIC(ext)* \cite{zeng2020cropping}& MobileNetV2 & GAICD        & 0.773 & 54.0  & 73.0   \\ 
        GAIC(ext) \cite{zeng2020cropping} & VGG16       & GAICD        & 0.741 & 53.3  & 69.6   \\
        CGS \cite{li2020composing}        & VGG16       & GAICD        & 0.773 & 54.7  & 72.0   \\ \hline
        Ours(basic)                       & VGG16       & GAICD        & 0.744 & 52.0  & 70.5   \\ 
        Ours                              & VGG16       & GAICD        & \textbf{0.795} & \textbf{59.7}  & \textbf{77.0}   \\
        \hline
        \end{tabular}
    \end{center}
\end{table*}

\begin{table*}[tbp]
    \caption{Comparison with the state-of-the-art methods on human-centric images in FCDB \cite{chen2017quantitative} and FLMS \cite{fang2014automatic} datasets. 
    GAIC(ext) \cite{zeng2020cropping} is the extension of GAIC\cite{zeng2019reliable}. The results marked with * are obtained using the released models from original papers}
    \label{table:comparison_FCDB}
    \begin{center}
        \setlength{\tabcolsep}{3mm}
        \begin{tabular}{l|cc|cc}
        \hline
        Method                              & Backbone      & Training Data & IoU$\uparrow$   & Disp$\downarrow$  \\ \hline \hline
        VFN* \cite{chen2017learning}         & AlextNet      & Flickr       & 0.5114 & 0.1257 \\
        VFN \cite{chen2017learning}         & VGG16         & CPC          & 0.6509 & 0.0876 \\
        VEN* \cite{wei2018good}              & VGG16         & CPC          & 0.6194 & 0.0930  \\
        VEN \cite{wei2018good}              & VGG16         & CPC          & 0.6670  & 0.0837 \\
        ASM-Net \cite{Tu2020ImageCW}       & VGG16         & CPC          & 0.7084 & 0.0755 \\
        LVRN* \cite{lu2019listwise}         & VGG16         & CPC          & 0.7373 & 0.0674 \\
        GAIC(ext)* \cite{zeng2020cropping}   & MobileNetV2   & GAICD        & 0.7126 & 0.0724 \\
        GAIC(ext) \cite{zeng2020cropping}   & VGG16         & CPC          & 0.7260 & 0.0708 \\
        CGS \cite{li2020composing}         & VGG16         & CPC          & 0.7331 & 0.0689 \\
        CACNet \cite{hong2021composing}    & VGG16         & FCDB         & 0.7364 & 0.0676 \\ \hline
        Ours(basic)                       & VGG16         & CPC          & 0.7263 & 0.0695 \\
        Ours                            & VGG16         & CPC          & \textbf{0.7469} & \textbf{0.0648} \\ \hline
        \end{tabular}
    \end{center}
\end{table*}

\noindent\textit{Content-preserving Feature:} Based on row 1, we add our content-preserving feature and report the results in row 7, in which we concatenate content-preserving feature with the region feature extracted from basic feature map. The results show the effectiveness of content-preserving feature. Next, we conduct ablation studies (row 8-11) based on row 7. 
We first replace GCN with conventional convolution layers (row 8) and observe the performance drop, which proves that it is useful to exploit the mutual relation between different regions. 
Then, we remove the predicted heatmap $\widetilde{\mathbf{H}}$ (row 9), resulting in significant performance drop, which highlights the importance of important content information. 
Additionally, we replace the proposed pseudo ground-truth heatmap with the saliency map detected by \cite{goferman2012context} in row 10 and obtain inferior performance. As discussed in Section \ref{sec:content_preserve}, this can be attributed to that saliency may not accurately reflect the content of interest for human-centric images.     
We also try using the content-preserving feature alone. Specifically, we only use content-preserving feature $\mathbf{f}^c$ to predict the aesthetic score (row 11). The performance is even worse than row 1, because the content-preserving feature is lacking in detailed content information and thus insufficient for aesthetic prediction.

\subsection{Comparison with the State-of-the-arts}
\label{sec:comparison}
\noindent\textit{Quantitative comparison:} We compare the performance of our model with the state-of-the-art methods on 50 human-centric images of GAICD \cite{zeng2019reliable} dataset in Table \ref{table:comparison_GAICD}. For the baselines with released models, we evaluate their models on the test set and report the results (marked with *). However, their backbone and training data may be different from our setting.

For fair comparison, we use the pretrained VGG16 \cite{vgg-simonyan2014very} as the backbone for all baselines and train them on GAICD dataset, based on their released code or our own implementation.
For our method, we additionally report the results of a basic version (``Ours(basic)") without using partition-aware feature or content-preserving feature (row 1 in Table \ref{table:ablation_study}).
It can be seen that Ours(basic) yields similar results with GAIC(ext) because they adopt the same region feature extractor (RoI+RoD). Among the baselines, GAIC(ext)* \cite{zeng2020cropping} and CGS \cite{li2020composing} are two competitive ones, owning to the more advanced architecture and the exploitation of mutual relations between different crops. Finally, our model outperforms all the state-of-the-art methods, which demonstrates that our method is more well-tailored for the human-centric image cropping task.

Apart from GAICD dataset \cite{zeng2019reliable}, we also collect 176 and 39 human-centric images from existing FCDB \cite{chen2017quantitative} and FLMS \cite{fang2014automatic} datasets, respectively, and compare our method with the state-of-the-art methods on these two datasets in Table \ref{table:comparison_FCDB}. 
Following \cite{wei2018good,Tu2020ImageCW}, we train the model on CPC dataset \cite{wei2018good}, and use IoU and Disp as evaluation metrics. 
Additionally, we adopt the strategy in \cite{lu2019listwise} to generate candidate crops and return the top-1 result as best crop without extra post-processing for all methods except CACNet \cite{hong2021composing}, which is trained to regress the best crop directly. 
As shown in Table \ref{table:comparison_FCDB}, our proposed model produces better results, but the performance gain is less significant than that on GAICD dataset. As claimed in \cite{zeng2019reliable}, one possible reason is that the IoU based metrics used in FCDB and FLMS datasets are not very reliable for evaluating cropping performance.
Furthermore, we also evaluate our method on both human-centric and non-human-centric images, and present results in Supplementary. 

\begin{figure*}[t]
    \centering
    \includegraphics[width=1\linewidth]{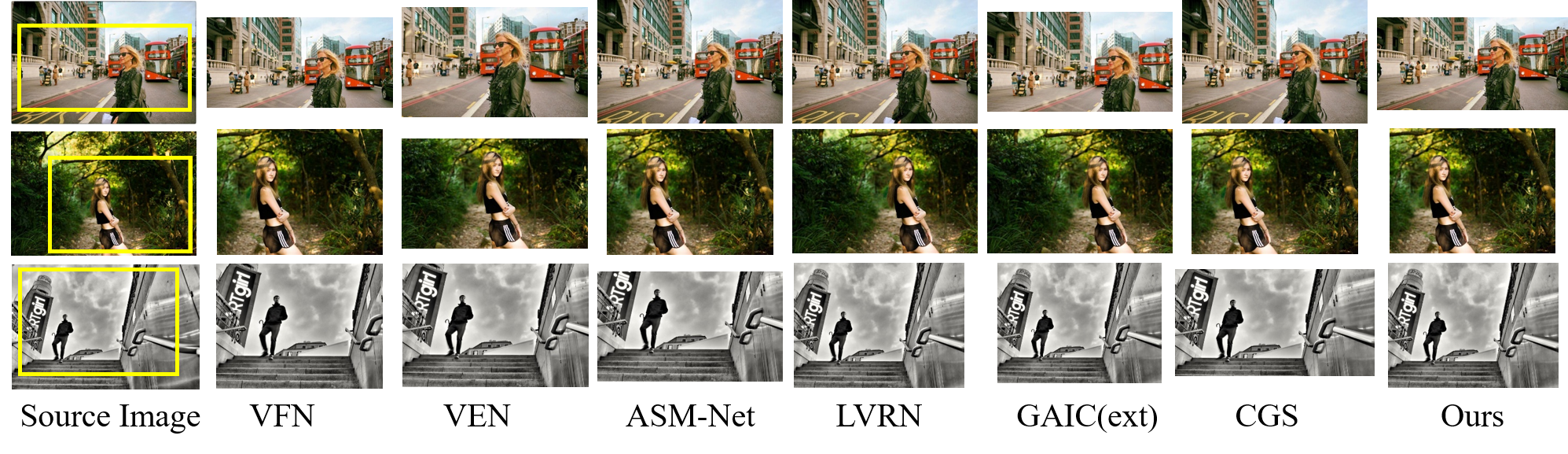}
    \caption{Qualitative comparison of different methods on human-centric images. We show the best crops predicted by different methods, which demonstrate that our method can generate better results close to the ground-truth best crops (yellow)}
    \label{fig:qualitative_comparison}
\end{figure*}

\begin{figure}[t]
    \centering
    \includegraphics[width=0.8\linewidth]{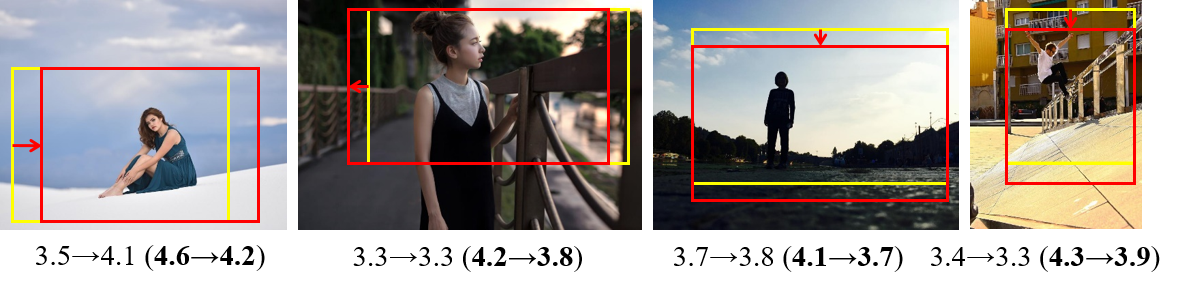}
    \caption{Examples of partition-aware feature enhancing the discrimination power of basic feature. Given the ground-truth best crop of an image (yellow) and another crop with similar content (red), we show their scores predicted by using basic feature (out of bracket) and partition-aware feature (in bracket), respectively}
    \label{fig:move_crop}
\end{figure}

\noindent\textit{Qualitative comparison:} We further conduct qualitative comparison to demonstrate the ability of our model in Figure \ref{fig:qualitative_comparison}. For each input image, we show the source image and the returned best crops by different methods, which demonstrates that our method can perform more reliable content preservation and removal. For example, in the first row of Figure \ref{fig:qualitative_comparison}, our method preserves more content on the left of human, probably because the person walks right to left, and reduces the top area that may hurt the image composition quality. In the second row, given the opposite face orientations to the first row, our model performs obviously different content preservation on the left/right sides of the human, yielding visually appealing crop.
More qualitative results are shown in Supplementary.

\subsection{Analysis of the Partition-aware Feature}
\label{sec:analysis_partition_feature}
To take a close look at the impact of partition-aware feature on candidate crop evaluation, we use the region features extracted from basic feature map and partition-aware feature map to predict scores for crops, respectively, corresponding to row 1 and row 2 in Table \ref{table:ablation_study}. 
As shown in Figure \ref{fig:move_crop}, to ensure that crop pairs have different aesthetic value yet similar content, for each image, we generate crop pair by moving its ground-truth best crop horizontally or vertically, in which the new crop still contains the human subject.
We can see that using partition-aware feature consistently leads to larger and more reasonable score changes than basic feature despite the various face orientations or postures of the human in Figure \ref{fig:move_crop}, which is beneficial for correctly ranking crops with similar content.

\begin{figure}[t]
    \centering
    \includegraphics[width=0.7\linewidth]{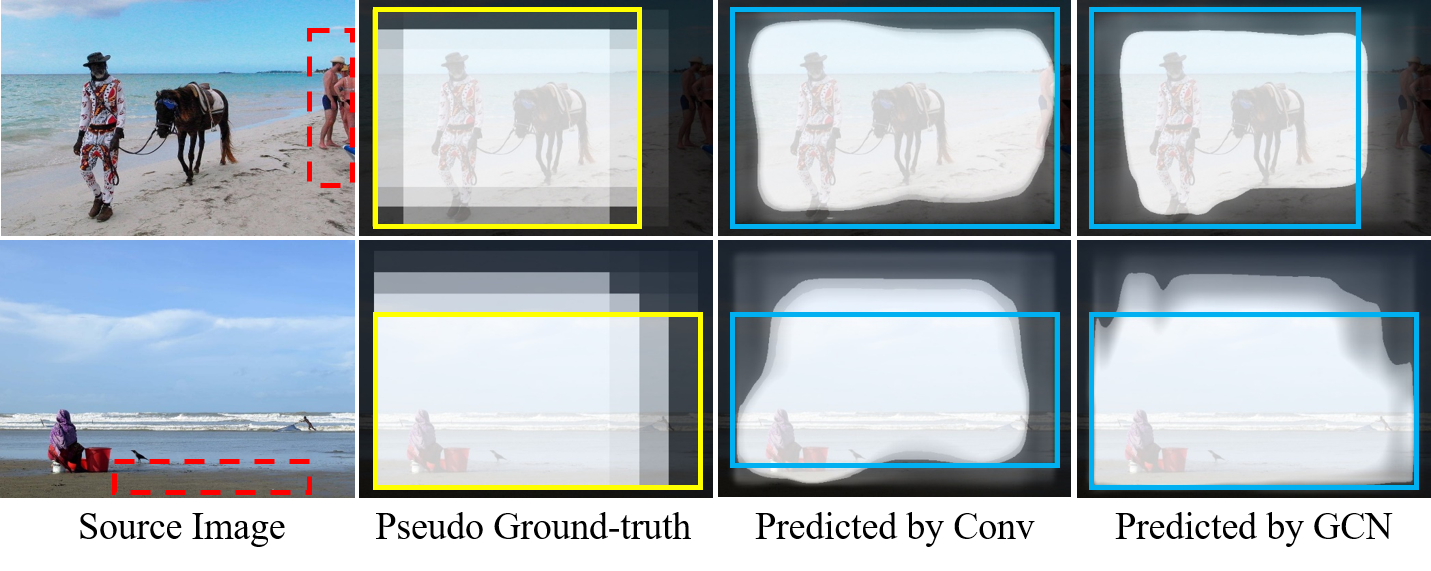}
    \caption{Visualization of the heatmap that indicates aesthetically important content. We show the source image, its pseudo ground-truth heatmap, the heatmaps estimated by conventional convolution(``Conv'') / graph convolution (``GCN''). We also draw the ground-truth (\emph{resp.}, predicted) best crops in yellow (\emph{resp.}, blue) boxes}
    \label{fig:heatmap_analysis}
\end{figure}

\subsection{Analysis of the Heatmap}
\label{sec:analysis_heatmap}
The ablation study in Section \ref{sec:ablation_study} demonstrates the superiority of graph-based relation mining (``GCN") over the conventional convolution (``Conv") when predicting the heatmap of important content  (see row 7,8 in Table \ref{table:ablation_study}). To reveal their difference qualitatively, we show the source image, its pseudo ground-truth heatmap, the heatmap predicted by ``conv"/``graph" convolution in Figure \ref{fig:heatmap_analysis}. With GCN learning the mutual relation between different regions, the model can make a more reasonable estimation of important content, especially the border area. For example, in the source image in the first row, we show an unpleasant outer area in the red dashed box, which should be removed for composing a good crop. The unimportant content (low values in the heatmap) predicted by ``GCN" completely covers the unpleasant area, while ``Conv" only covers part of the unpleasant area. In the second row, unlike ``Conv" that only deems the area around person as important, ``GCN" predicts relatively high values for the area behind person, indicating that preserving such area in a crop may be beneficial. 
In summary, ``GCN" can facilitate important content localization and contributes to more informative content-preserving feature.

\section{User Study}
\label{sec:user_study}
Given the subjectiveness of aesthetic assessment task, we conduct user study to compare different methods, in which we employ total 265 human-centric images, 176 from FCDB \cite{chen2017quantitative}, 50 from GAICD \cite{zeng2019reliable}, and 39 from FLMS \cite{fang2014automatic}. For each image, we generate 7 best crops by using seven different methods: VFN \cite{chen2017learning}, VEN \cite{wei2018good}, ASM-Net \cite{Tu2020ImageCW}, LVRN \cite {lu2019listwise}, GAIC(ext) \cite{zeng2020cropping}, CGS \cite{li2020composing}, and our proposed method. 20 experts are invited to select the best result for each image. 
Then we calculate the percentage that the results generated by different methods are selected as the best ones. The percentages of the abovementioned six baselines are 1.7\%, 5.6\%, 9.0\%, 14.6\%, 15.7\%, and 22.5\%, respectively, while our method achieves the highest percentage 30.9\% and clearly outperforms the other methods. 

\begin{figure}[t]
    \centering
    \includegraphics[width=0.8\linewidth]{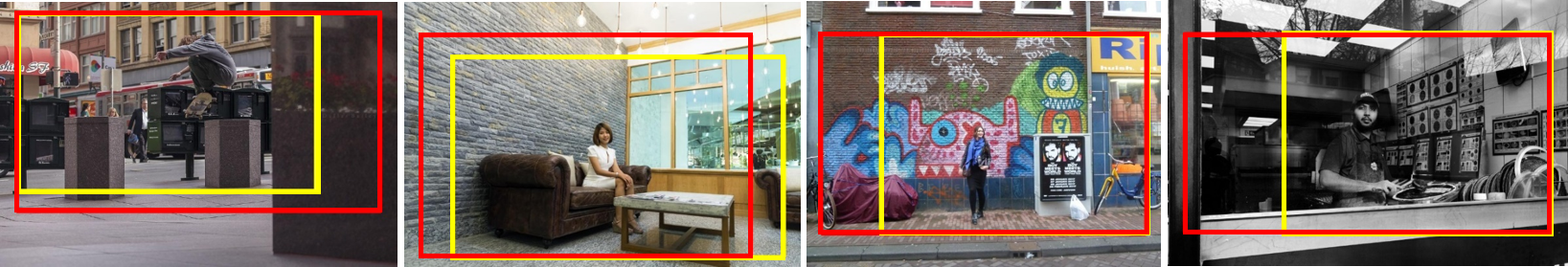}
    \caption{Some failure cases in the test set of GAICD dataset \cite{zeng2019reliable}. For each image, the ground-truth and predicted best crops are drawn in yellow and red boxes, respectively}
    \label{fig:failure_cases}
\end{figure}
\section{Limitations}
\label{sec:limitations}
Our method can generally produce reliable crops for human-centric images, but it still has some failure cases. Some failure cases in the test set of GAICD dataset \cite{zeng2019reliable} are shown in Figure \ref{fig:failure_cases}, where the best crops produced by our method are far away from the ground-truth one and rank relatively low in the ground-truth annotations. For these examples, our method tends to preserve similar areas on the left/right side of the human subject in the best crop, probably because the complicated backgrounds and confusing human information (\eg, inconsistent orientations between face and body) compromise the effectiveness of partition-aware feature and content-preserving feature.

\section{Conclusion}
In this paper, we have proposed a human-centric image cropping method with novel partition-aware and content-preserving features. The partition-aware feature allows to treat different regions in a candidate crop differently conditioned on the human information. The content-preserving feature represents how well each candidate crop preserves the important content.
Extensive experiments have demonstrated that the proposed method can achieve the best performance on human-centric image cropping task.

\section*{Acknowledgement}
The work is supported by Shanghai Municipal Science and Technology Key Project (Grant No. 20511100300), Shanghai Municipal Science and Technology Major Project, China (2021SHZDZX0102), and National Science Foundation of China (Grant No. 61902247).

\clearpage
%
%
\bibliographystyle{splncs04}
\bibliography{egbib}
\end{document}


\pagestyle{headings}
\mainmatter
\def\ECCVSubNumber{4249}  

\title{Supplementary for Human-centric Image Cropping with Partition-aware and Content-preserving Features} 

\titlerunning{Human-centric Image Cropping}
%
\author{Bo Zhang\orcidlink{0000-0001-6577-7484} \and
Li Niu\thanks{Corresponding author}\orcidlink{0000-0003-1970-8634} \and
Xing Zhao\orcidlink{0000-0001-9234-3235} \and
Liqing Zhang\orcidlink{0000-0001-7597-8503}}
\authorrunning{B{.} Zhang et al.}
%
\institute{MoE Key Lab of Artificial Intelligence, Shanghai Jiao Tong University, China \\ \email{\{bo-zhang,ustcnewly,1033874657\}@sjtu.edu.cn, zhang-lq@cs.sjtu.edu.cn}
}
\maketitle

In this document, we provide additional materials to supplement our main submission. We first present more implementation details in Section \ref{sec:implement_details} and describe how to determine the main subject for human-centric image in Section \ref{sec:main_subject}. Next, we report the performance of our model using inaccurate human bounding box in Section \ref{sec:performance_inaccurate} and evaluate two simplest baselines that use human bounding box to directly produce cropping box in Section \ref{sec:simplest_methods}.
In Section \ref{sec:hyper_analysis}, we study the impact of using different hyper-parameters in our method. Then, we present the results of evaluating our method on general image cropping and compare with the state-of-the-arts in Section \ref{sec:quantitative_results}, and compare the running speed and model complexity of different models in Section \ref{Sec:speed_complexity}. 
Moreover, we show more qualitative results in Section \ref{sec:qualitative_results}. 

\section{Implementation Details}
\label{sec:implement_details}
We conduct experiments on Ubuntu 18.04 equipped with an NVIDIA RTX 3090 GPU and 64GB RAM. We use the Adam optimizer~\cite{kingma2015adam} with a weight decay of $1e^{-4}$ to train the model for 80 epochs. 
We randomly select 64 crops for one image when training on GAICD dataset \cite{zeng2019reliable}, while we use all the crops of an image when training on CPC dataset \cite{wei2018good}.
Following \cite{li2020composing}, we use the warmup \cite{he2016deep} in the first five epochs to increase the learning rate from 0 to $1e^{-4}$ and decay the learning rate with a cosine annealing \cite{loshchilov2016sgdr} in the following epochs. 

\section{Main Subject Determination}
\label{sec:main_subject}
Our method is proposed to handle images with a single person and considers the other people as background when multiple subjects exist. In our implementation, given multiple human objects in an image, we take the person that appears in the ground-truth best crop of the image as the main subject.  
In real-world applications, we can adopt a simple approach to determine the main subject, by selecting the object with a larger area and closer to the image center through weighted ranking. Specifically, the rank score of a human object is defined as: $w*h + \alpha \big(1 - \sqrt {\left( {x - 0.5 } \right)^2 + \left( {y - 0.5 } \right)^2 }\big)$ and we take the object with largest rank score as the main subject. Here, given a human bounding box, $w$ and $h$ are its width and height normalized by the size of source image, respectively, and $(x,y)$ indicates the normalized coordinates of the bounding box center. The weight $\alpha$ is set as 0.1 empirically.
To verify the utility of this simple approach, we first select all images with more than one detected human objects from GAICD \cite{zeng2019reliable} and CPC \cite{wei2018good} datasets, in which we obtain 60 and 216 samples, respectively. 
Then we use above approach to determine the main subject and compare it with the manually selected results. 
As a result, this approach achieves an accuracy of 98.0\% on GAICD dataset and 97.6\% on CPC dataset, demonstrating its utility in determining the main subject for human-centric images.

\section{Performance using Inaccurate Human Bounding Box}
\label{sec:performance_inaccurate}
Recall the proposed partition-aware feature depends on the human bounding box detected by Faster R-CNN~\cite{ren2015faster} trained on Visual Genome~\cite{krishna2017visual}, which can generally provide reliable human detection results.
Nevertheless, the predicted human bounding box may still be inaccurate or even wrong. 
For quantitative study on the performance of human detection, we first define a detected bounding box, whose intersection over union (IoU) with the manually checked human bounding box is below 0.5, as a missed detection case. Note that we only keep at most one human bounding box per image and determine the main subject using the approach mentioned in Section \ref{sec:main_subject} when detecting more than one human objects in an image. 
Meanwhile, a human-centric image without person detected is also regarded as a missed detection case and we treat it as a non-human-centric image to generate crops (see Section 3.4 of the main text). 

In such case, we find that the detected human bounding boxes have a missed detection rate of 3.4\% in the test sets of GAICD \cite{zeng2019reliable}, FCDB \cite{chen2017quantitative}, and FLMS \cite{fang2014automatic} datasets, which have a total of 265 human-centric images.
Then we evaluate the proposed model using the above inaccurate human bounding box, yielding the results: $\overline{SRCC}$=0.790, $\overline{Acc_5}$=59.3, $\overline{Acc_{10}}$=76.8 on GAICD dataset, $IoU$=0.745, $Disp$=0.0658 on FCDB and FLMS datasets, which are comparable with the performance of the proposed method using manually checked human bounding boxes (see Table 2 and Table 3 of the main text).
This is probably because we perform partition on the feature map (with a large receptive field), rather than input image, which supports the model to be more robust to the human bounding boxes. 

\section{Comparison with the Simplest Baselines}
\label{sec:simplest_methods}
As described in Section 3.2 of the main text, we use the human bounding box to derive partition-aware feature from the basic feature and apply partition-aware feature to help improve human-centric image cropping.
Apart from the proposed method, naturally, there are some other approaches to leverage human bounding box for human-centric image cropping.
We take two simplest ways as baselines: directly take the human bounding box as cropping box (denoted by ``Baseline\_A'') and select a crop that contains a larger human area and places the person closer to the crop center from the pre-defined candidate crops through weighted ranking (denoted by ``Baseline\_B''). 
Specifically, given the human bounding box $B_h$ and a candidate crop $B_c$, ``Baseline\_B'' calculates the rank scores of this candidate as: $IoU(B_h,B_c) + \sqrt{2} - Dist(B_h,B_c)$, in which $IoU(B_h,B_c)$ is the intersection over union between the candaite crop and human bounding box: $area(B_h \cap B_c) / area(B_h \cup B_c)$, and $Dist(B_h,B_c)$ denotes the euclidean distance between their centers normalized by the size of source image. After calculating rank scores of all candidates, ``Baseline\_B'' takes the candidate crop with largest rank score as the best crop.

Due to only producing one cropping box, these two baselines cannot be evaluated using the evaluation metrics of GAICD dataset \cite{zeng2019reliable}, \ie, $\overline{SRCC}$, $\overline{Acc_5}$, and $\overline{Acc_{10}}$. So we evaluate them on human-centric images in FCDB \cite{chen2017quantitative} and FLMS \cite{fang2014automatic} datasets using IoU and Disp as evaluation metrics, whose results are reported in Table \ref{table:comparison_baselines}.
\begin{table*}[t]
    \caption{Comparison with the simplest baselines on human-centric images in FCDB \cite{chen2017quantitative} and FLMS \cite{fang2014automatic} datasets. Baseline\_A takes the human bounding box as output. Baseline\_B selects a crop that contains a larger human area and places the person closer to the center from pre-defined candidates}
    \label{table:comparison_baselines}
    \begin{center}
        \setlength{\tabcolsep}{3mm}
        \begin{tabular}{l|cc|cc}
        \hline
        Method                              & Backbone      & Training Data & IoU$\uparrow$   & Disp$\downarrow$  \\ \hline \hline
        Baseline\_A                         & -             & -             & 0.3634 & 0.1600 \\
        Baseline\_B                         & -             & -             & 0.3903 & 0.1541 \\ \hline
        Ours(basic)                         & VGG16         & CPC           & 0.7263 & 0.0695 \\
        Ours                                & VGG16         & CPC           & \textbf{0.7469} & \textbf{0.0648} \\ \hline
        \end{tabular}
    \end{center}
\end{table*}
For comparison, in Table \ref{table:comparison_baselines}, we also display the performance of our method and its basic version (``Ours(basic)"). We can see that those simplest baselines perform significantly worse than our basic model and are also inferior to existing state-of-the-arts (see Table 3 of the main text), which may attribute to the plain equal treatment of all regions outside the human for human-centric image cropping. In contrast, by using partition-aware, our model can treat different regions in a crop differently conditioned on the human information. 

\section{Hyper-parameter Analysis}
\label{sec:hyper_analysis}
\begin{figure}[t]
    \centering
    \includegraphics[width=0.55\linewidth]{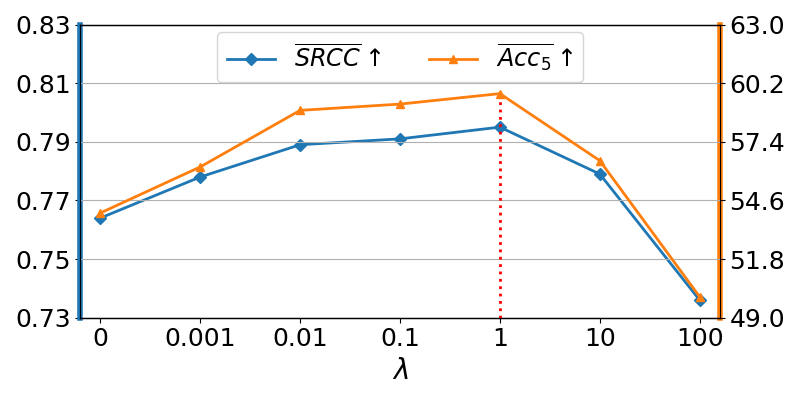}
    \caption{Performance variation of our method with different trade-off parameter $\lambda$ in Eqn.(7) of the main text on GAICD dataset. The dashed vertical line denotes the default value used in other experiments}
    \label{fig:hyper_exp}
\end{figure}

Recall that we have a trade-off parameter $\lambda$ in Eqn.(7) of the main text, which is set as $1$ via cross-validation by splitting 20\% training samples as validation set. We report the results on test set in Figure~\ref{fig:hyper_exp} when $\lambda$ varies from  0 to 100, using the average Spearman's rank-order correlation coefficient ($\overline{SRCC}$) and averaged top-5 accuracy ($\overline{Acc_{5}}$). Comparing the result without content loss ($\lambda=0$) and the result with $\lambda=1$, we can see a clear gap between their performance. When $\lambda=0$, the quality of predicted heatmap cannot be guaranteed without the supervision of pseudo ground-truth and the low-quality heatmap may harm the performance. When $\lambda$ becomes larger than 1, the performance begins to drop. The experimental results in Figure~\ref{fig:hyper_exp} demonstrate that our model is robust when setting $\lambda$ in the range of [0.01,1].
\section{Evaluation on General Image Cropping}
\label{sec:quantitative_results}
In Section 4.4 of the main text, we have shown that the proposed method outperforms existing state-of-the-art methods on the human-centric images of several benchmarks.
Recall our method can perform image cropping on both human-centric and non-human-centric images (see Section 3.4 of the main text), and the proposed content-preserving feature is also useful for general image cropping, because preserving important content is also crucial for non-human-centric images. So we evaluate our model on the whole test set of GAICD dataset \cite{zeng2019reliable}, which contains 50 human-centric images and 150 non-human-centric images. In Table \ref{table:comparison_GAICD}, we compare our method with different state-of-the-arts on GAICD and use three metrics for performance evaluation: the average Spearman's rank-order correlation coefficient ($\overline{SRCC}$) and averaged top-$N$ accuracy ($\overline{Acc_{N}}$) for both $N=5$ and $N=10$.
For our method, we additionally report the results of a basic version (``Ours(basic)") without using partition-aware feature or content-preserving feature.
We can see that Ours(basic) yields similar results with GAIC(ext) \cite{zeng2020cropping}, which attributes to that they adopt the same region feature extractor (RoI+RoD). Ours outperforms Ours(basic) significantly, which demonstrates the effectiveness of content-preserving feature for the general image cropping task.
Among these methods, CGS \cite{li2020composing} is the most competitive one, probably because it exploits mutual relations between different crops.
\begin{table*}[t]
    \caption{Quantitative comparison on the whole test set of GAICD dataset \cite{zeng2019reliable}. GAIC(ext) \cite{zeng2020cropping} is the extension of GAIC\cite{zeng2019reliable}. The results of other methods are from original papers, except for the results of VFN and VEN are from \cite{zeng2019reliable}. Two best results for each metric are highlighted in boldface}
    \label{table:comparison_GAICD}
    \begin{center}
        \setlength{\tabcolsep}{2mm}
        \begin{tabular}{l|cc|ccc}
        \hline
        Method                          & Backbone    & Training Data   & $\overline{SRCC}\uparrow$   & $\overline{Acc_5}\uparrow$ & $\overline{Acc_{10}}\uparrow$ \\ 
        \hline \hline
        VFN \cite{chen2017learning}     & AlextNet    & Flickr          & 0.450                       & 26.7                       & 38.7   \\
        VEN \cite{wei2018good}          & VGG16       & CPC             & 0.621                       & 37.6                       & 50.9   \\
        ASM-Net \cite{Tu2020ImageCW}    & VGG16       & GAICD           & 0.766                       & 54.3                       & 71.5   \\
        GAIC(ext) \cite{zeng2020cropping}& MobileNetV2 & GAICD          & 0.783                       & 57.2                       & \textbf{75.5}   \\
        CGS \cite{li2020composing}      & VGG16       & GAICD           & \textbf{0.795}              & \textbf{59.7}              & \textbf{77.8}   \\ 
        \hline
        Ours(basic)                     & VGG16       & GAICD           & 0.777                       & 54.3                       & 71.0   \\
        Ours                            & VGG16       & GAICD           & \textbf{0.793}           & \textbf{58.6}           & 74.5   \\
        \hline
        \end{tabular}
    \end{center}
\end{table*}

Despite that the proposed approach outperforms the state-of-the-art methods including CGS \cite{li2020composing} when evaluating on human-centric images, it does not achieve the best performance on the whole test set of GAICD. 
One possible reason is that non-human-centric image cropping cannot benefit from the proposed partition-aware feature, while there are significantly more non-human-centric images than human-centric ones in the test set of GAICD with a ratio of 3:1.
The other reason may lie in the generation mechanism of candidate crops in GAICD \cite{zeng2019reliable}, which constrains the area of candidate crops to preserve the major content of the source image.
Recall the content-preserving feature is proposed to augment our method by learning how well each candidate crop preserves the important content. For the images that place the important content in the center, the performance gain of the proposed content-preserving feature would be limited by the aforementioned mechanism.
In practice, the area of good crops is varying and the candidate crops should be sampled across different sizes and positions.

Apart from GAICD dataset, we also evaluate on the whole FLMS dataset \cite{zeng2019reliable}, which has total 500 images including 39 human-centric images, and present results in Table \ref{table:comparison_FLMS}, in which we train the model on CPC dataset \cite{wei2018good}, and use intersection of union (IoU) and boundary displacement (Disp) as evaluation metrics following \cite{wei2018good}. Ours outperforms Ours(basic) significantly, which again proves that our modification is also beneficial for general image cropping.
Among these baselines, CACNet \cite{hong2021composing} achieves the best performance, probably because it is trained to directly regress the best crop without using predefined candidate crops. Then, our model also performs comparably with the strongest CACNet in terms of IoU and Disp. Moreover, the proposed method can perform favorably against CGS \cite{li2020composing} on FLMS, which verifies the utility of content-preserving feature. 

In summary, although the focus of this work is human-centric image cropping, our method can still achieve competitive performance on the general image cropping task.

\begin{table}[t]
    \caption{Quantitative comparison on the whole FLMS dataset \cite{zeng2019reliable}. All methods use VGG16 \cite{vgg-simonyan2014very} as backbone. The results of other methods are from original papers. To measure the efficiency, we also report FPS and model complexity of different models. Two best results for each metric are highlighted in boldface}
    \label{table:comparison_FLMS}
    \begin{center}
        \setlength{\tabcolsep}{2.5mm}
        \begin{tabular}{l|c|cc|ccc}
        \hline
        Method                          &Training  & IoU$\uparrow$     & Disp$\downarrow$   
        &FPS$\uparrow$    &FLOPs  &Parameters\\ 
        \hline \hline
        VEN\cite{wei2018good}           &CPC            & 0.837             & 0.041 
        &10     &15.39G &40.93M\\
        ASM-Net\cite{Tu2020ImageCW}     &CPC            & 0.849             & 0.039 
        &102    &64.36G &14.95M\\
        LVRN\cite {lu2019listwise}      &CPC            & 0.843             & -     
        &270    &15.39G &40.93M\\
        GAIC\cite{zeng2019reliable}     &GAICD          & 0.834             & 0.041 
        &\textbf{299}    &20.07G &16.31M\\
        CGS\cite{li2020composing}       &GAICD          & 0.836             & 0.039 
        &174    &20.08G &21.25M\\
        CAC-Net\cite{hong2021composing} &FCDB           & \textbf{0.854}    & \textbf{0.033} 
        &\textbf{323}   &16.26G &18.93M\\
        \hline
        Ours(basic)                     &CPC            & 0.832             & 0.042 
        &211    &20.17G &17.90M\\
        Ours                            &CPC            & \textbf{0.850} & \textbf{0.034} 
        &128    &20.25G &19.47M\\
        \hline
        \end{tabular}
    \end{center}
\end{table}

\section{Running Speed and Model Complexity}
\label{Sec:speed_complexity}
A practical image cropping model should have fast speed and acceptable computational complexity for real-time implementation.
We compare the running speed in terms of frame-per-second (FPS) and model complexity of our method with different state-of-the-arts, and report results in the last three columns of Table \ref{table:comparison_FLMS}. 
All models are tested on the same PC with i9-10920X CPU, 64G RAM and one NVIDIA RTX 3090 GPU, and our method as well as other methods that do not provide codes is implemented under the Pytorch toolbox.
Following previous works \cite{zeng2019reliable,li2020composing}, the running speed of all methods is evaluated on GAICD dataset \cite{zeng2019reliable}, in which each image has about 86 candidate crops, and our method is tested on the human-centric images of GAICD dataset using both partition-aware and content-preserving features.
Moreover, the reported FLOPs of different models are calculated on image with the resolution used in their original papers. 

There are three points worth noting: 1) VEN \cite{wei2018good} and LVRN \cite {lu2019listwise} adopt the same network architecture, leading to the same FLOPs and number of parameters. 
2) The high FPS of CAC-Net \cite{hong2021composing} can be attributed to its one-stage regression manner, which is significantly different from the general pipeline used in our method.
3) For the methods that do not provide codes, including ASM-Net \cite{Tu2020ImageCW}, CGS \cite{li2020composing}, and CAC-Net \cite{hong2021composing}, their FLOPs and number of parameters are calculated based on our implementations, which may be different from them reported in their original papers due to different implementation details.   
In Table \ref{table:comparison_FLMS}, we can see that the proposed method runs slower than some state-of-the-art methods yet still at 128 FPS, which enables our model to be applied to practical applications with real-time implementation requirement.
Regarding the model complexity, the FLOPs (the number of multiply-adds) and number of parameters of our model are more than the most efficient GAIC \cite{zeng2019reliable} and CAC-Net \cite{hong2021composing}, but comparable with most other competitors.
Furthermore, the comparison between our model and its basic version (``Ours(basic)'') indicates the low computational resources of additional modules, \ie, 80M FLOPs and 1.57M parameters, which is almost ignorable compared to the computational cost of our basic model.

\begin{figure*}[tbp]
    \centering
    \includegraphics[width=1\linewidth]{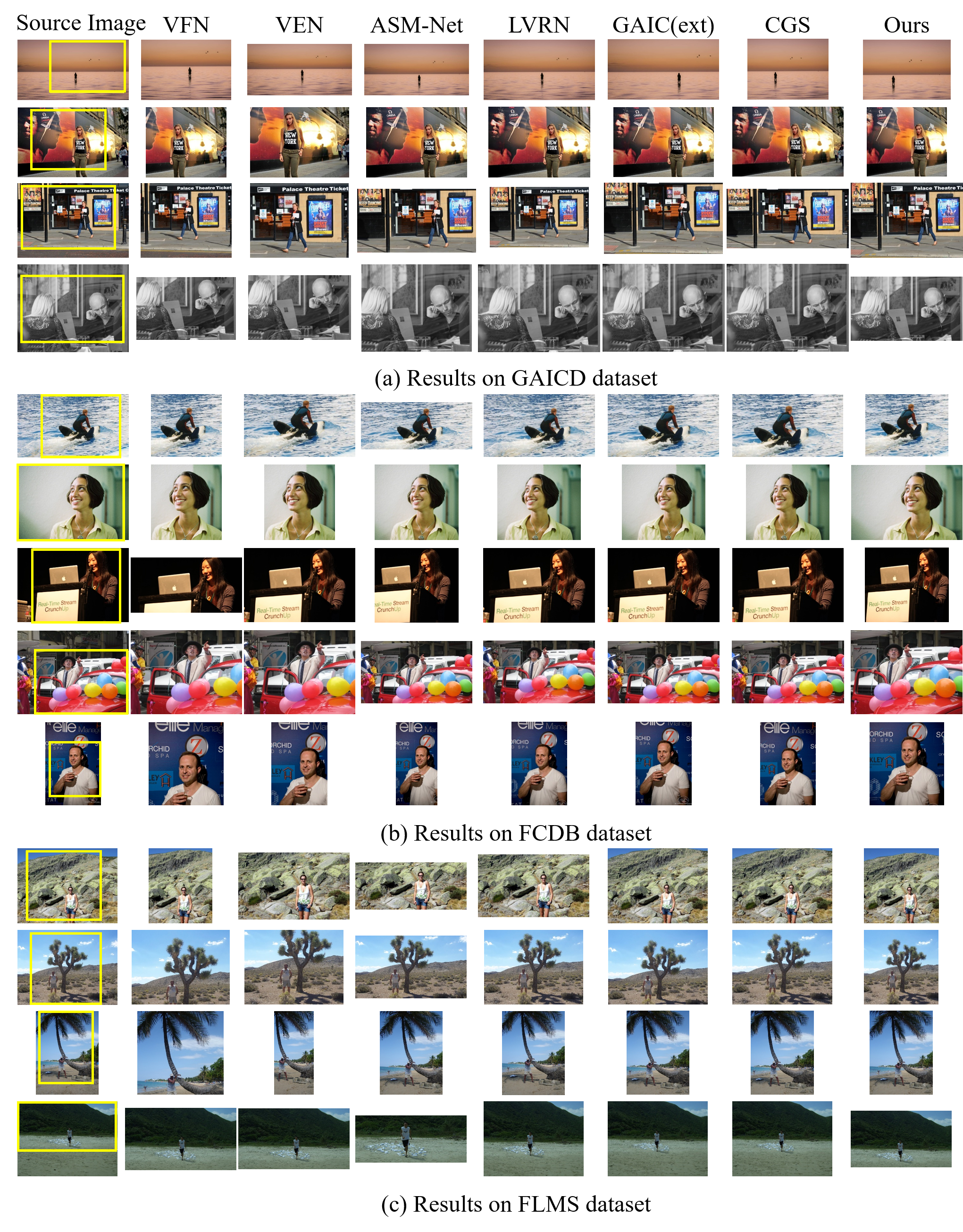}
    \caption{Qualitative comparison of different methods on the human-centric images of GAICD \cite{zeng2019reliable}, FCDB \cite{chen2017quantitative}, and FLMS \cite{fang2014automatic} datasets. We show the best crops predicted by different methods, demonstrating that our method can generate better results close to the ground-truth best crops (yellow)}
    \label{fig:more_qualitative_comparison}
\end{figure*}

\section{More Qualitative Results}
\label{sec:qualitative_results}
To better demonstrate the effectiveness of our method, we present more qualitative comparisons with the state-of-the-art methods on the human-centric images of GAICD \cite{zeng2019reliable},  FCDB \cite{chen2017quantitative}, and FLMS \cite{fang2014automatic} datasets.
In Figure \ref{fig:more_qualitative_comparison}, we show the source image, the ground-truth best crop, and the predicted best crops by different methods. For the images in FLMS dataset that are assigned with more than one ground-truth crop, we randomly select one to draw on the image.

It can be seen that the proposed method generally produces more appealing cropping results close to the ground truth, which attributes to more reasonable content preservation and removal. For example, in the second row on GAICD dataset, most compared approaches cannot entirely remove the extraneous content on the right side, while the proposed method works well. In the second row on FCDB dataset, considering that the person looks at the upper right, it is supposed to preserve more corresponding area in the returned best crop. However, most methods fail on it, but our method accomplishes it correctly. Those examples further validate the utility of our method for human-centric image cropping.  

\begin{figure*}[tbp]
    \centering
    \includegraphics[width=1\linewidth]{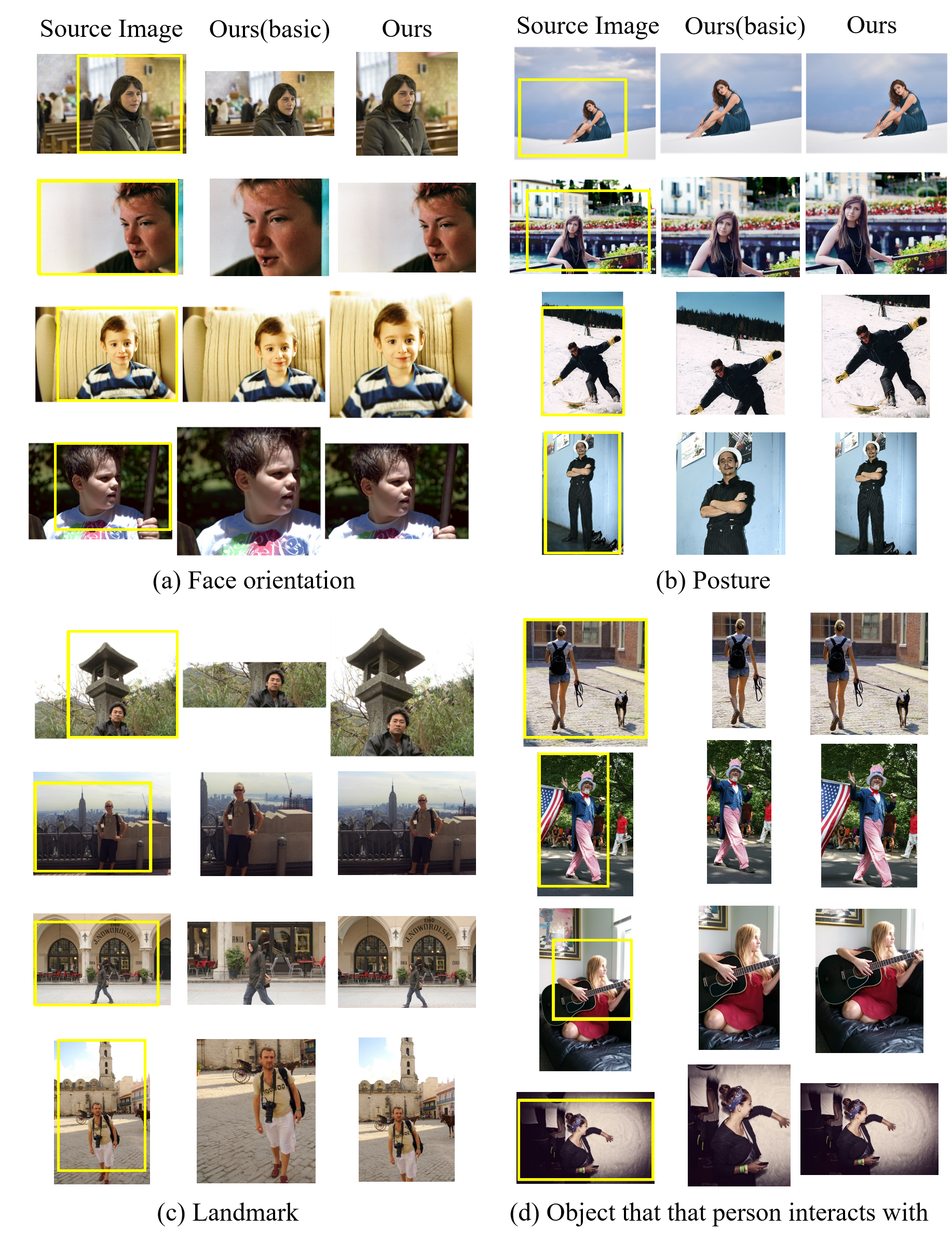}
    \caption{Qualitative comparison of our model and its basic version without using partition-aware feature or content-preserving feature. The examples are divided into four groups according to the concerned content (under each subfigure), \eg, group (a) evaluates model on the images with various face orientations. The yellow box indicates the ground-truth best crop}
    \label{fig:different_human}
\end{figure*}

Additionally, to take a close look at the superiority of the proposed method, we compare the cropping results of our method and its basic version (``Ours(basic)") without using partition-aware feature or content-preserving feature in Figure \ref{fig:different_human}. 
To study the sensitivity to human of our model, we show the cropping results of our/basic models on the images with different face orientations and postures in Figure \ref{fig:different_human} (a) and (b), respectively, from which we see that the proposed model can adaptively generate well-composed cropping according to the human information and generally produce more reliable crops than the basic model. 
For example, in the first two rows of Figure \ref{fig:different_human} (a), when the person looks to the right (\resp, left), our model preserves more content on the right (\resp, left) of human, yielding visually pleasing crops. Meanwhile, in the third row, when the person looking straight ahead, our model places the human in the crop center and removes the distracting contents, leading to good composition with visual balance. However, for above images, the basic model performs inflexible and typically places the human closer to the crop center regardless of different face orientations, resulting in inferior cropping results.
Similarly, in Figure \ref{fig:different_human} (b), the proposed model is capable of incorporating human posture and performing flexible content preservation/removal for the area around person, which can generate more appealing crops than the basic model. 
Their performance differences can be attributed to the partition-aware feature, which enables our model to treat different regions in a crop differently conditioned on the human information.

Besides human information, we also evaluate our/basic models on the images containing the important content that are often ignored by saliency detection methods, such as interesting objects (\eg, landmark) and the objects that person interacts with.
In Figure \ref{fig:different_human} (c), we show the results on the images with landmark behind the person. It can be seen that the our model can roughly capture the entire landmark in the predicted crops, while the basic model may fail to include the landmark very well. 
We conjecture that this is primarily driven by the important content estimation adopted in the proposed model, which is supervised by the pseudo heatmap that highlights the attractive objects (\eg, landmark) in the background and the objects that person interacts with (see Section 3.3 of the main text).
In Figure \ref{fig:different_human} (d), compared with the basic model, our model can preserve these objects that person interacts with (\eg, dog, flag, and guitar) better, which benefits from the aforementioned pseudo heatmap and graph-based region relation mining that exploits the mutual relation between different regions for important content estimation (see Section 4.6 of the main text).
Those qualitative comparisons further confirm the effectiveness of the proposed partition-aware and content-preserving features on human-centric image cropping task.


%
%
\bibliographystyle{splncs04}
\bibliography{egbib}